\documentclass[letterpaper, 10 pt, conference]{ieeeconf}  

\IEEEoverridecommandlockouts                              
                                          
\usepackage{graphicx} 
\usepackage{times} 
\usepackage{amsmath} 
\usepackage{amssymb}  
\usepackage{bm}
\usepackage[caption=false,font=footnotesize]{subfig}
\usepackage{float}
\usepackage{hyperref}

\graphicspath{{./figures/}}

\title{\LARGE \bf
Nonlinear Controller Design for Quadrotor with Inverted Pendulum
}

\author{Xichen Shi and Yashwanth Kumar Nakka
\thanks{*This work was done as part of a course project. The authors contributed equally.}
\thanks{Authors are with Caltech at the time of the work. Email: {xshi, ynakka}@caltech.edu}
}%

\begin{document}

\maketitle
\thispagestyle{empty}
\pagestyle{empty}

\begin{abstract}
The quadrotor is a $6$ degrees-of-freedom (DoF) system with underactuation. Adding a spherical pendulum on top of a quadrotor further complicates the task of achieving any output tracking while stabilizing the rest. In this report, we present different types of controllers for the nonlinear dynamical system of quadrotor and pendulum combination, utilizing feedback-linearization and control Lyapunov function with quadratic programming (CLF-QP) approaches. We demonstrated trajectory tracking for quadrotor-only case as well as quadrotor-pendulum-combined case.
\end{abstract}

\section{INTRODUCTION}
The inverted pendulum is a classical nonlinear control problem that has been widely used as a fundamental system for testing several control algorithms. 
Its interest has been motivated by many applications, e.g. to stabilize walking pattern of bipedal robots, in which legs are modeled as double pendulum, and of self-balancing wheeled robots.
Advanced control strategies, which are able to balance it at the upright position, have been investigated by many researchers.
In \cite{sarkar2017application}, a Linear Quadratic Regulator (LQR) controllers and Model Reference Adaptive Controllers (MRAC) have been successfully implemented for the swinging-up control problem of the pendulum.
The classical control problem of the inverted pendulum can be extended by placing it on top of an aerial vehicle such as a quadrotor. The flying inverted pendulum is a nonlinear, underactuated,and inertially coupled system with $8$ DoFs ($6$ from the quadrotor and $2$ from the pendulum) and $4$ control inputs. The unstable zero dynamics of the coupled system for a non zero initial perturbation of the pendulum makes trajectory tracking in all the DoFs difficult to achieve. Here, we design nonlinear controllers using feedback-linearization and control Lyapunov function with quadratic programming to demonstrate position trajectory tracking with altitude and attitude as the output variables. In the following, we present the dynamic model of an inverted pendulum that is inertially coupled to a quadrotor. The flying inverted pendulum has an unstable equilibrium, which is represented by the upwards vertical position of the pendulum. The analysis presented in the following is aimed at stabilizing the pendulum while simultaneously tracking a position trajectory. 

The paper is organized as follows. In the section~\ref{sec:model}, we present a detailed model of a quadrotor with an inverted pendulum. Control design for position trajectory tracking of quadrotor-only and the coupled quadrotor-pendulum system is described in section~\ref{sec:control_design_quad}. The trajectory tracking was demonstrated using Matlab simulations as presented in section~\ref{sec:simulation_results}. We conclude the paper with possible extensions of the proposed controllers.

\section{MODEL}
\label{sec:model}
\subsection{Quadrotor Dynamics}
Quadrotor states are defined as position: $\mathbf{p} = [p_X, p_Y, p_Z]^\top$; velocity: $\mathbf{v} = [v_X, v_Y, v_Z]^\top$; Euler angles: $\mathbf{q} = [\phi, \theta, \psi]^\top$; and body angular velocity: $\bm{\omega} = [\omega_x, \omega_y, \omega_z]^\top$. Furthermore, it is assumed to have a mass of $m$, and a diagonal inertia matrix $I = \mathrm{diag}[I_{xx}, I_{yy}, I_{zz}]$. The dynamics are given by:
\begin{align}
    \ddot{\mathbf{p}} &= \mathbf{g} + \mathbf{g}_1(\mathbf{q})f_z \\
    \dot{\mathbf{q}} &= Z(\mathbf{q})\bm{\omega} \\
    \bm{\dot{\omega}} &= I^{-1}(I\bm{\omega} \times \bm{\omega}) + I^{-1}\bm{\tau}
\end{align}
where $\mathbf{g}_1(\mathbf{q})$ and $Z(\mathbf{q})$ are defined as
\begin{align*}
    \mathbf{g}_1(\mathbf{q}) &=
    \begin{bmatrix}
    -\frac{1}{m}(\sin{\phi}\sin{\psi} + \cos{\phi}\sin{\theta}\cos{\psi})\\
    -\frac{1}{m}(-\sin{\phi}\cos{\psi} + \cos{\phi}\sin{\theta}\sin{\psi})\\
    -\frac{1}{m}\cos{\phi}\cos{\theta}
    \end{bmatrix} \\
    Z(\mathbf{q}) &= 
    \begin{bmatrix}
    1 & \sin{\phi}\tan{\theta} & \cos{\phi}\tan{\theta} \\
    0 & \cos{\phi} & -\sin{\phi} \\
    0 & \sin{\phi}\sec{\theta} & \cos{\phi}\sec{\theta}
    \end{bmatrix}
\end{align*}
$[f_z, \bm{\tau}]^\top = [f_z, \tau_x, \tau_y, \tau_z]^\top$ are the body-z force and three axis moments generated by four rotors. Suppose we can control individual rotor speed $\mathbf{u} = [u_1, u_2, u_3, u_4]^\top$ directly, then the relationship can be represented as
\begin{equation}
    \begin{bmatrix}
    f_z \\ \tau_x \\ \tau_y \\ \tau_z
    \end{bmatrix}
    =
    \underbrace{
    \rho D^4
    \begin{bmatrix}
    C_T & C_T & C_T & C_T \\
    0 & C_T \cdot l & 0 & -C_T \cdot l \\
    C_T \cdot l & 0 & C_T \cdot l & 0 \\
    C_Q & C_Q & C_Q & C_Q 
    \end{bmatrix}
    }_{B}
    \begin{bmatrix}
    u_1 \\ u_2 \\ u_3 \\ u_4
    \end{bmatrix}
\end{equation}

\subsection{Pendulum Dynamics}
\begin{figure}[!t]
    \centering
    \includegraphics[width=0.4\textwidth]{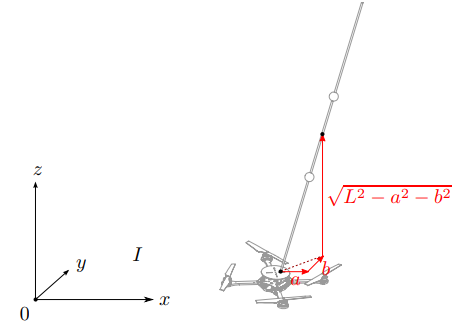}
    \caption{Quadrotor-pendulum coordinate system~\cite{brescianini2013quadrocopter}}
    \label{fig:quad_pend}
\end{figure}
A spherical pendulum having mass of $m_p$ and length $2L$ with 2 DoFs is attached to the center-of-mass (CoM) of the quadrotor. The CoM of pendulum is parametrized as global coordinate offset from $\mathbf{p}$ by $[a,b,\zeta]^\top$, shown in Fig.~\ref{fig:quad_pend}, and based on geometric constraint $\zeta = \sqrt{L^2 - a^2 - b^2}$. In addition, it is assumed that the mass of pendulum is much smaller compared to the quadrotor, thus the latter dynamics is unaffected. Dynamics of the pendulum can be obtained through Lagrangian formulation~\cite{brescianini2013quadrocopter}:
\begin{equation}
\begin{bmatrix}\ddot{a} \\ \ddot{b} \end{bmatrix} = \mathbf{f}_p(a,b,L,\dot{a},\dot{b},\zeta) + B_p(a,b,L)\ddot{\mathbf{p}}
\label{eq:pend_dyn}
\end{equation}
with 
\begin{align}
    \mathbf{f}_p(a,b,L,\dot{a},\dot{b},\zeta) &= \begin{bmatrix} \frac{a H(a,b,\dot{a},\dot{b})}{4L^2\zeta^2}\\ \frac{b H(a,b,\dot{a},\dot{b})}{4L^2\zeta^2}\end{bmatrix} \\
    B_p(a,b,L) &= 
    \begin{bmatrix} \frac{3(a^2 - L^2)}{4L^2} & \frac{3ab}{4L^2} & \frac{3a\zeta}{4L^2} \\
    \frac{3ab}{4L^2} & \frac{3(b^2 - L^2)}{4L^2} &\frac{3b\zeta}{4L^2}\end{bmatrix}   
\end{align}
where 
\begin{equation*}
    H(a,b,\dot{a},\dot{b})= 4\dot{b}^2(a^2-L^2) - 8\dot{a}\dot{b}ab + 4\dot{a}^2(b^2-L^2) + 3\zeta^3g
\end{equation*}
The form of equation in~\eqref{eq:pend_dyn} is to facilitates control system design which will become obvious later.

\section{CONTROL SYSTEM DESIGN}
\label{sec:control_design_quad}
The control objective is to ensure that the quadrotor tracks a time dependent position trajectory with  specified performance bounds. Two control strategies were used to design the trajectory tracking controller: 1) Feedback linearization and 2) Control Lyapunov Function as a Quadratic Program (CLF-QP). In this section, we first describe the outputs considered for feedback linearization and use consequent normal form to design the controller using CLF-QP. 

\subsection{Control Design for Quadrotor (Output Tracking)}
\label{subsec:Quad_Control}
\subsubsection{Output}
We chose to work with $\mathbf{y} = [p_Z, \phi, \theta, \psi]^\top$ as outputs. Suppose we would like to do set-point regulation on output $\mathbf{y} = [p_Z, \phi, \theta, \psi]^\top$ to desired $\mathbf{y}_d$.
Differentiate $\mathbf{y}$ until we can get the form
\begin{equation}
     \mathbf{y}^{\bm{\gamma}} = L_f^*h + A(x)B \mathbf{u}
     \label{eq:fbl_y}
\end{equation}
with
\begin{equation}
    L_f^*h = \begin{bmatrix}
    g \\
    \cdot \\
    \frac{\partial\big(Z(\mathbf{q})\bm\omega\big)}{\partial\mathbf{q}} Z(\mathbf{q})\bm\omega + Z(\mathbf{q})I^{-1}(I\bm{\omega} \times \bm{\omega})\\
    \cdot
    \end{bmatrix}
\end{equation}


\begin{equation}
    A(x) = \begin{bmatrix}
    -\frac{1}{m}\cos{\phi}\cos{\theta} & 0 \\
    0 & Z(\mathbf{q})I^{-1}
    \end{bmatrix}
\end{equation}
which indicates a vector relative degree $\bm{\gamma} = [2,2,2,2]$. 

\subsubsection{Feedback Linearization}
From~\eqref{eq:fbl_y}, we can design feedback linearizing controller as
\begin{equation}
    \mathbf{u} = \left( A(x)B\right)^{-1}(-L_f^*h + \mathbf{v})
\end{equation}
Then the closed-loop output dynamics becomes
\begin{equation}
    \bm{\dot{\eta}} = F \bm{\eta} + G \mathbf{v}
\end{equation}
with $\bm{\eta} = [(\mathbf{y}-\mathbf{y}_d)^\top, \dot{\mathbf{y}}^\top]^\top$, and $F$,$G$ defined as
\begin{align}
    F &=
    \begin{bmatrix}
    0_{4 \times 4} & \mathcal{I}_{4 \times 4} \\
    0_{4 \times 4} & 0_{4 \times 4} 
    \end{bmatrix} \\
    G &= 
    \begin{bmatrix}
    0_{4 \times 4}  \\
    \mathcal{I}_{4 \times 4}
    \end{bmatrix}
\end{align}
Naturally, we can pick any positive definite $Q$ and solves the Continuous-time Algebraic Ricatti Equation (CARE): $F^\top P + PF - P G G^\top P + Q = 0$ and set $\mathbf{v} = - G^\top P \bm{\eta}$. Substitute into $\mathbf{u}$:
\begin{equation}
    \mathbf{u} = (A(x)B)^{-1}(-L_f^*h - G^\top P \bm{\eta})
    \label{eq:feedback-lin}
\end{equation}

\subsubsection{Output Trajectory Tracking using Feedback Linearization}
Consider the following output dynamics, as described earlier. To track a time dependent trajectory $\mathbf{y}_{d}(t)$ the feedback linearizing controller is designed as in Eq.~(\ref{eq:feedback-lin-tracking}). 
\begin{align}
     \mathbf{y}^{\bm{\gamma}} &= L_f^*h + A(x)B \mathbf{u}
\end{align}
\begin{equation}
    \mathbf{u} = (A(x)B)^{-1}\left(-L_f^*h + \ddot{\mathbf{y}}_{d} - \alpha_2 (\dot{\mathbf{y}} - \dot{\mathbf{y}}_{d}) - \alpha_1 (\mathbf{y} - \mathbf{y}_{d}) \right)
    \label{eq:feedback-lin-tracking}
\end{equation}

\subsubsection{Position Trajectory Tracking using Feedback Linearization}
\label{subsubsec:force-allocation}
The horizontal position dynamics $\mathbf{p}_{x},\mathbf{p}_{y}$ forms the internal dynamics of the quadrotor when feedback linearization is done with the outputs considered in this paper. The position tracking is performed by computing Euler angles required to apply desired acceleration in the $x,y$ direction. Equation~(\ref{eq:posit-trak-force}) presents the equations used to compute the desired Euler angles to track a given position trajectory $\mathbf{p}$. 

\begin{align}
\ddot{\mathbf{p}} &= \mathbf{g} + \mathbf{g}_1(\mathbf{q})f_z\\ 
& = \ddot{\mathbf{p}}_d + K (\dot{\mathbf{p}}_d - \dot{\mathbf{p}}) + K(\mathbf{p}_d - \mathbf{p})
\label{eq:posit-trak-force}
\end{align}

In the following equation, $\mathbf{f}_d = [f_{xd}, f_{yd}, f_{zd}]^\top$ where $\mathbf{f}_{d} = \mathbf{g}_1(\mathbf{q})f_z $. 
\begin{subequations}
\begin{align}
f_{xd} &= \ddot{x}_d + K_d (\dot{x}_d - \dot{x}) + K_p(x_d - x)\\
f_{yd} &= \ddot{y}_d + K_d (\dot{y}_d - \dot{y}) + K_p(y_d - y)\\
f_{zd} &= -g + \ddot{z}_d + K_p (\dot{z}_d - \dot{z}) + K_p(z_d - z)
\end{align}\label{eq:force-allocation}
\end{subequations}

The desired Euler angles given in Eq.~(\ref{eq:desired_att}) are computed by assuming yaw angle $\psi_d = 0$.

\begin{subequations}
\begin{align}
    \psi_d   &= 0 \\
    \phi_d   &= \arcsin\big(\frac{-f_{xd}\sin\psi_d + f_{yd}\cos\psi_d}{\lVert \mathbf{f}_d \rVert}\big) \\
    \theta_d &= \arctan\big(\frac{f_{xd}\cos\psi_d + f_{yd}\sin\psi_d}{f_{zd}}\big);
\end{align}
\label{eq:desired_att}
\end{subequations}
Using the desired Euler angles computed using the desired position trajectory, the output tracking controller is implemented to track the desired position trajectory.

\subsubsection{CLF-QP}
Thus the Lyapunov function $V(\bm{\eta}) = \bm{\eta}^\top P \bm{\eta}$ is a CLF and there exists controllers that satisfy the following inequality:
\begin{equation}
    \inf\{L_F V + L_G V \mathbf{v}\} \leq -\frac{\lambda_{\min}(Q)}{\lambda_{\max}(P)}V(\bm{\eta})
\end{equation}
The rotor speeds $\mathbf{u}$ has to be within some bounds $\mathbf{u}_{\min} \leq \mathbf{u} \leq \mathbf{u}_{\max}$. Since $v$ and $u$ are related by
\begin{equation}
    A(x)B\mathbf{u} + L_f^*h(x) = \mathbf{v}
\end{equation}
and setting $c_3 = \lambda_{\min}(Q)/ \lambda_{\max}(P) $, we can arrive at the following quadratic program:
\begin{equation}
\begin{aligned}
& \underset{\mathbf{v}}{\text{minimize}} & & \mathbf{v}^\top \mathbf{v} \\
& \text{subject to} & & -\big(2\bm{\eta}^\top P G \big) \mathbf{v} &\geq& \bm{\eta}^\top \big(F^\top P + P F + c_3 P \big)\bm{\eta} \\
& & & \qquad (A(x)B)^{-1}\mathbf{v} &\geq& (A(x)B)^{-1}L_f^*h + \mathbf{u}_{\min} \\
& & & \qquad (A(x)B)^{-1}\mathbf{v} &\leq& (A(x)B)^{-1}L_f^*h + \mathbf{u}_{\max}
\end{aligned}
\label{eq:clf-qp}
\end{equation}

\subsubsection{Position Trajectory Tracking using CLF-QP}
The output trajectory tracking using CLF-QP is achieved similar to the feedback linearized controller. In this case, the states $\mathbf{\eta}$  are replaced by the error state defined as $\mathbf{\eta}= \mathbf{y} - \mathbf{y}_{d}$. 
The Equations described in section~\ref{subsubsec:force-allocation} are used in conjunction with CLF-QP to perform the position trajectory tracking as described for feedback linearization.

\subsection{Output Tracking for Inverted Pendulum}
\subsubsection{Feedback Linearization}
\label{sec:fbklin_pend}
Taking $\mathbf{y}_p = [a, b]^\top$ as output of the system, we see that it can be viewed as having relative degree $\bm{\gamma} = [2,2]$ if we treat $\ddot{\mathbf{p}}$ as input. Thus we can transform equation~\eqref{eq:pend_dyn} to the following temporary form
\begin{equation}
\begin{bmatrix}\ddot{a} \\ \ddot{b} \end{bmatrix} = \mathbf{f}_p(a,b,L,\dot{a},\dot{b},\zeta) + B_p(a,b,L)\bm{\xi}
\label{eq:pend_ctrl1}
\end{equation}
Since $B_p$ is a $2\times 3$ matrix, we can design the following feedback linearizing trajectory tracking controller
\begin{align}
    \bm{\xi} &= B_p^{\dagger}(-\mathbf{f}_p + \bm{\nu})\\
    \bm{\nu} &= \ddot{\mathbf{y}}_{pd} - K_1(\dot{\mathbf{y}}_p-\dot{\mathbf{y}}_{pd}) - K_2(\mathbf{y}_p-\mathbf{y}_{pd})
\end{align}
It is then straightforward to set $\ddot{\mathbf{p}}_d = \bm{\xi}$. Then the problem becomes identical as the position trajectory tracking problem in Sect.~\ref{subsec:Quad_Control}. We can then use~\eqref{eq:desired_att} to get the desired attitude $\mathbf{q}_d$.

Alternatively, we can also only use $\ddot{p_X}$ and $\ddot{p_Y}$ as input to~\eqref{eq:pend_ctrl1} and get a different formulation
\begin{equation}
\begin{bmatrix}\ddot{a} \\ \ddot{b} \end{bmatrix} = \mathbf{f}^\prime_p(a,b,L,\dot{a},\dot{b},\zeta, \ddot{p_Z}) + B^\prime_p(a,b,L)\bm{\xi}^\prime
\label{eq:pend_ctrl2}
\end{equation}
and results in a slightly modified version of controller using inverse of $B^\prime_p$ instead of pseudo-inverse of $B_p$
\begin{equation}
    \bm{\xi}^\prime  = B_p^{\prime-1}(-\mathbf{f}_p + \bm{\nu})
\end{equation}

\subsubsection{Linear-Quadratic-Regulator}
\label{sec:lqr_pend}
Ideally, we would like to control both $\mathbf{p}$ and $\mathbf{y}_p$ to some degree. But due to the strong nonlinear coupling between the two from equation~\eqref{eq:pend_dyn}, traditional nonlinear controller synthesis techniques are not suitable. The problem would either be solved via numerical optimization off-line (trajectory planning) or online (Model predictive control)~\cite{brescianini2013quadrocopter,mellinger2011minimum}. However, for some less aggressive maneuver, the system can be stabilized on some nominal trajectory where linear dynamics are valid, and linearization around nominal trajectory or equilibrium can be useful in achieving the total output control~\cite{hehn2011flying}
\begin{equation}
    \begin{bmatrix}
    \ddot{a} \\
    \ddot{b} \\
    \ddot{p_X} \\
    \ddot{p_Y} \\
    \end{bmatrix}
    =
    \begin{bmatrix}
    (3g/4L)a + (3/4)g\theta \\
    (3g/4L)a - (3/4)g\phi \\
    -g\theta\\
    g\phi
    \end{bmatrix}
    \label{eq:pend_dyn_lin}
\end{equation}
By treating $\phi$ and $\theta$ as control inputs to the system, we can convert Eq.~\eqref{eq:pend_dyn_lin} to standard LTI form:
\begin{equation}
    \dot{\bm{\eta}}_p = A_p\bm{\eta} + B_p\begin{bmatrix}\phi_d\\ \theta_d\end{bmatrix}
\end{equation}
where $\bm{\eta}_p = [a,b,p_X,p_Y,\dot{a},\dot{b},\dot{p_X},\dot{p_Y}]$. We can show the system is fully controllable and thus enables us to design a full-state feedback LQR controller in the form of
\begin{equation}
    \begin{bmatrix}\phi_d\\ \theta_d\end{bmatrix}
    = -K_{\mathrm{lqr}}\bm{\eta}_p
\end{equation}
By assuming $\psi_d = 0$, we obtained the desired Euler angles and then employ previous controllers from Sect.~\ref{subsec:Quad_Control} to drive $\mathbf{q} \to \mathbf{q}_d $.

\section{SIMULATION RESULTS}
\label{sec:simulation_results}
\subsection{Quadrotor}
The controllers designed in the previous section are used to perform a trajectory where the quadrotor flies up to certain height $z = 2m$ and tracks a circular trajectory of radius 1m. It was observed that both the controllers perform the given task but differ in control effort and performance under process noise. 

\subsubsection{Feedback Linearization}
\begin{figure}[h!]
\centering{
\subfloat[$p_x$ tracking using feedback linearization]{
    \includegraphics[width=0.22\textwidth]{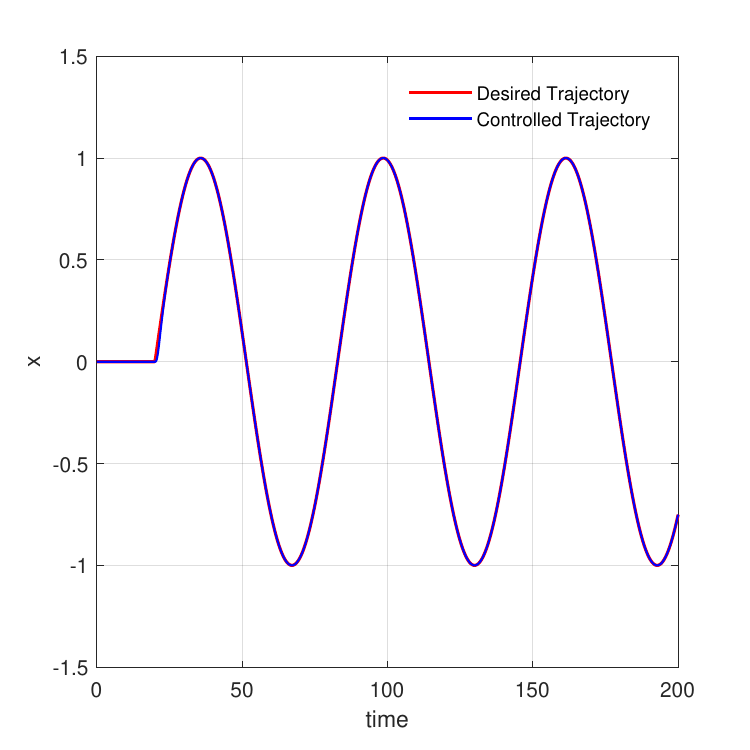}
    }
}
\centering{
\subfloat[$p_y$ tracking using feedback linearization]{
    \includegraphics[width=0.22\textwidth]{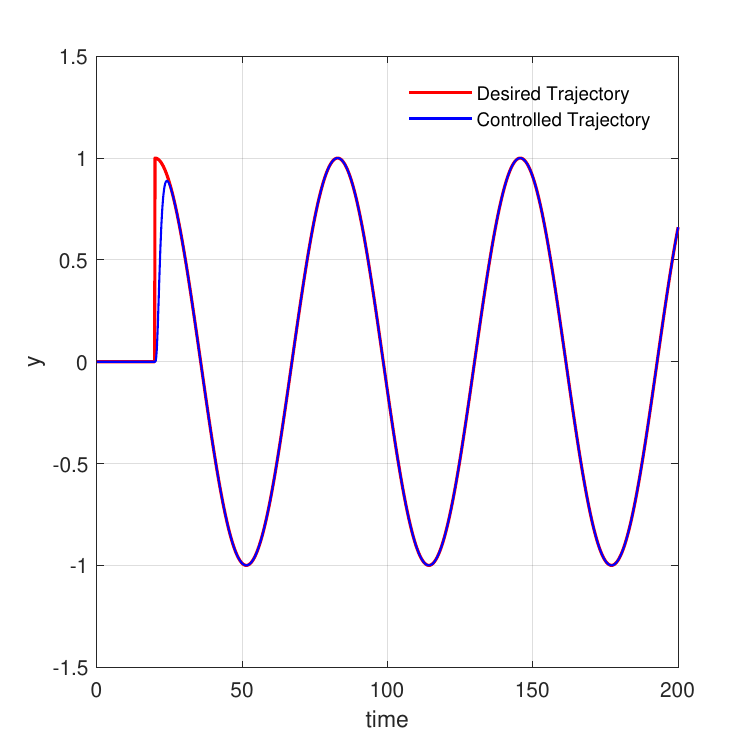}
    }
}
\centering{
\subfloat[$p_z$ stabilization using feedback linearization]{
    \includegraphics[width=0.22\textwidth]{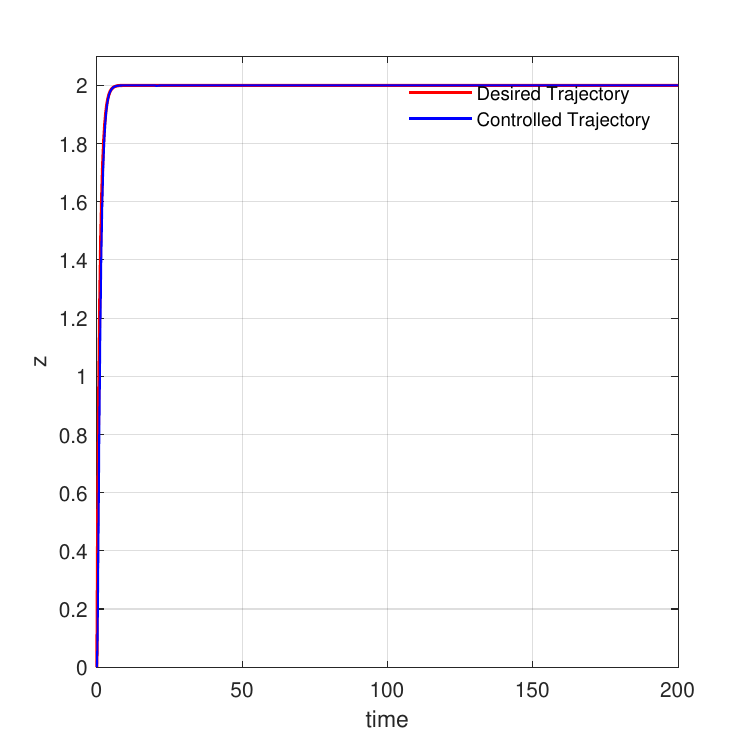}
    }
}
\centering{
\subfloat[3D trajectory tracking using feedback linearization]{
    \includegraphics[width=0.22\textwidth]{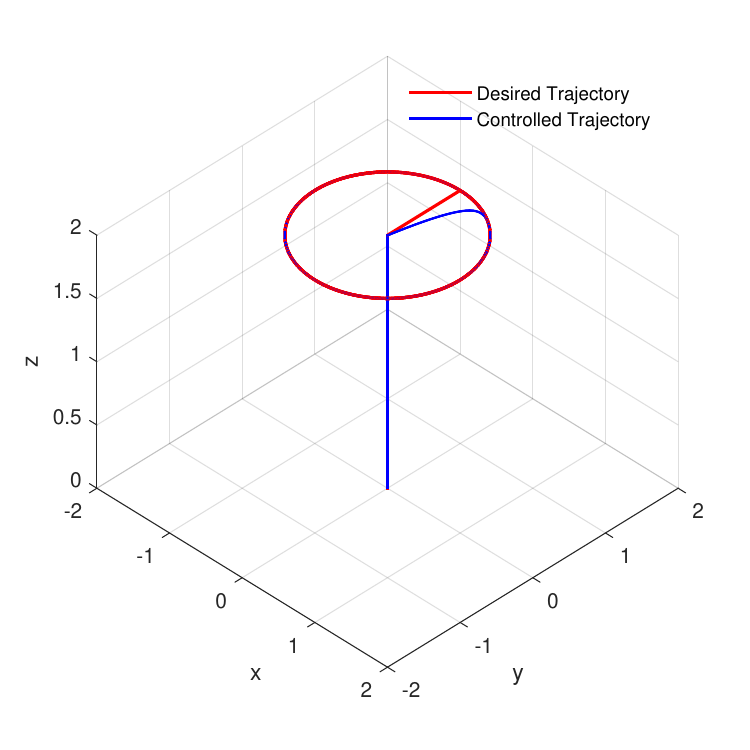}
    }
}
\caption{Trajectory tracking using feedback linearization in x,y direction while maintaining a constant altitude.}
\label{fig:FBL_tracking}
\end{figure}

The feedback linearization controller trajectory tracking results are shown in Fig.~\ref{fig:FBL_tracking}. The controller is tuned to have high damping to reduce over shoot and also to track the demanding harmonic trajectory in $x,y$ direction. It can be observed that the tracked trajectory is smoother than the desired trajectory at transition from flying up to just flying in the horizontal plane. The control effort plots are added in the shared box folder.

\subsubsection{CLF-QP}
\begin{figure}[H]
\centering{
\subfloat[$p_x$ tracking using CLF-QP]{
    \includegraphics[width=0.22\textwidth]{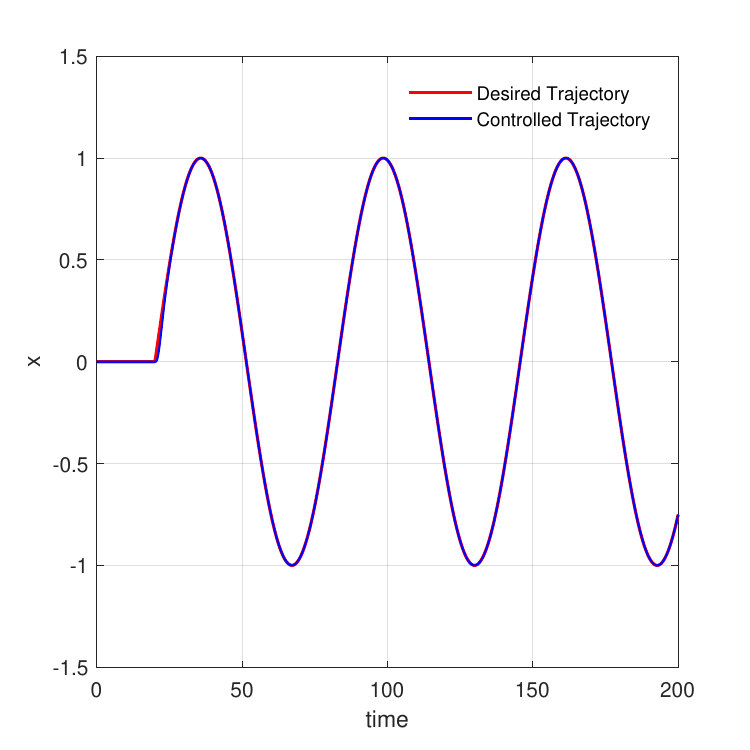}
    }
}
\centering{
\subfloat[$p_y$ tracking using CLF-QP]{
    \includegraphics[width=0.22\textwidth]{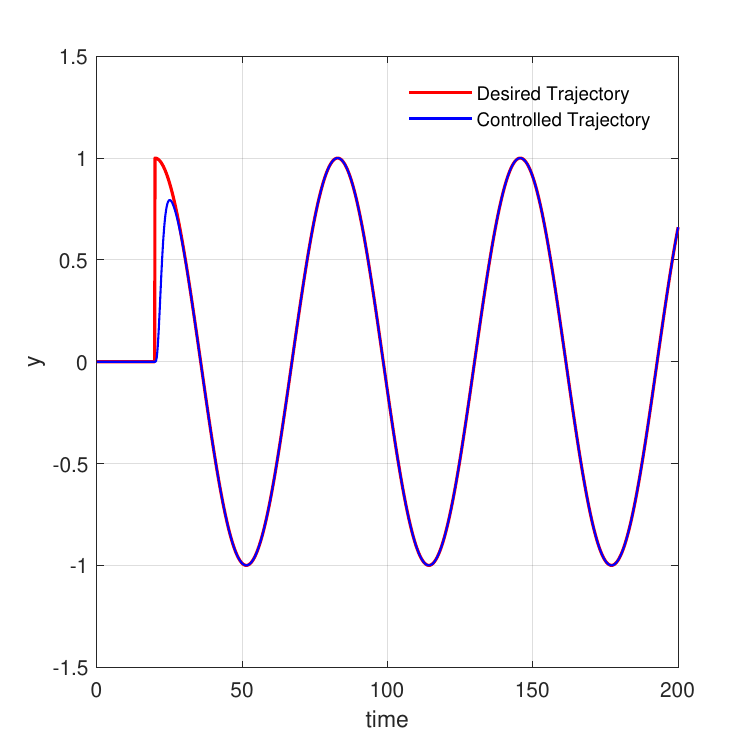}
    }
}
\centering{
\subfloat[$p_z$ stabilization using CLF-QP]{
    \includegraphics[width=0.22\textwidth]{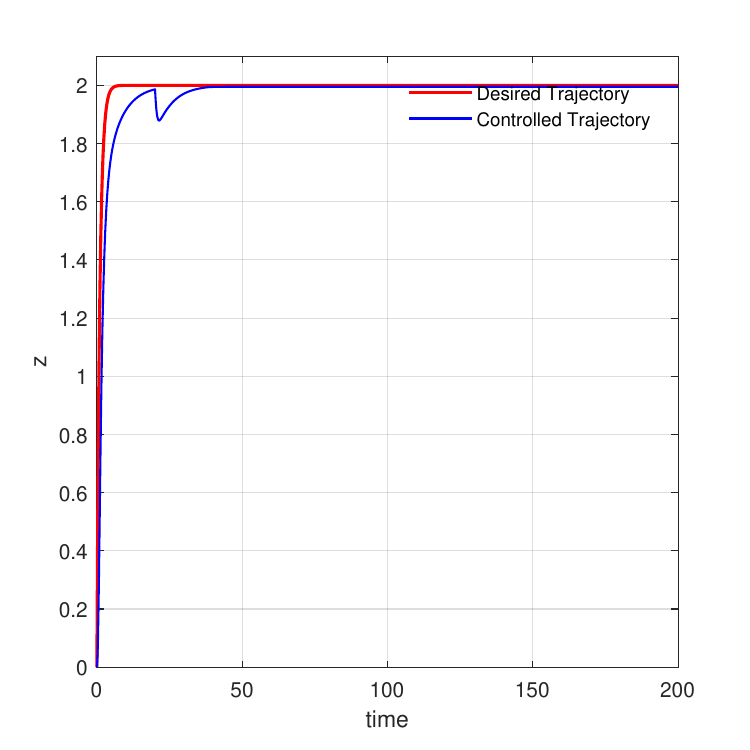}
    }
}
\centering{
\subfloat[3D trajectory tracking using CLF-QP]{
    \includegraphics[width=0.22\textwidth]{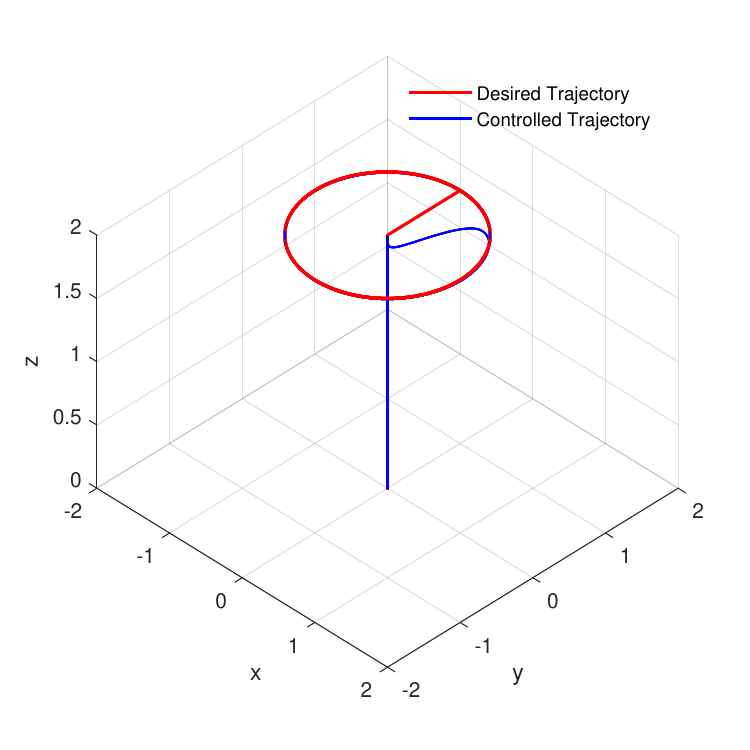}
    }
}
\caption{Trajectory tracking using CLF-QP in x,y direction while maintaining a constant altitude.}
\label{fig:CLF-QP-tracking}
\end{figure}

The CLF-QP controller performs better than feedback linearization control in steady state tracking. When the controller switches form hovering to moving in horizontal trajectory, slight loss of control in direction was observed to the high gradients in the desired trajectory which can be observed in Fig.~\ref{fig:CLF-QP-tracking}. The CLF-QP controller accommodates for the actuator saturation. It was observed that the $B$ matrix in cost function was giving a in-feasible solution, so it was included in the constraints. Further investigation is required to understand how control to rotor speed mapping is effecting the optimization result.   

\subsubsection{Tracking with process noise}
\begin{figure}[h!]
\centering{
\subfloat[$p_x$ tracking using feedback linearization]{
    \includegraphics[width=0.22\textwidth]{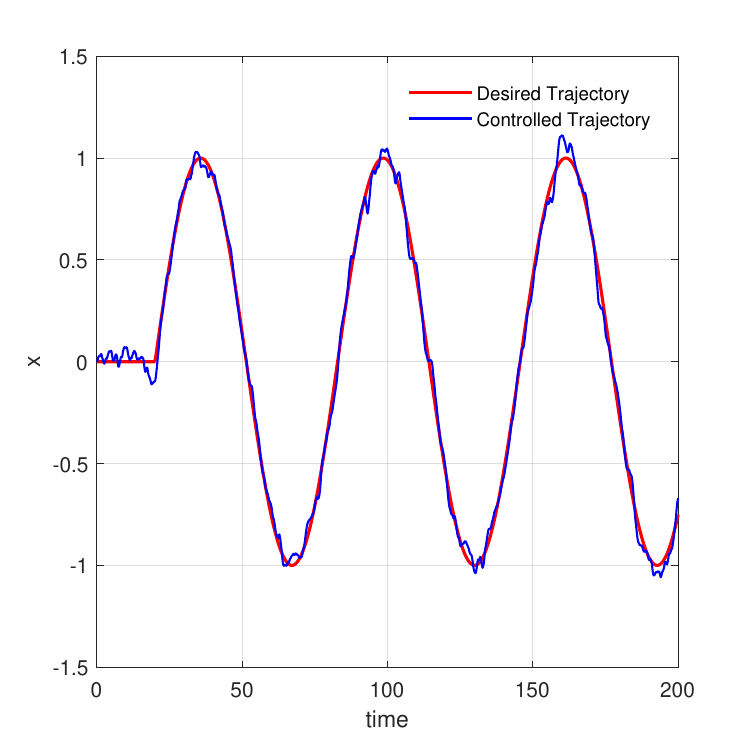}
    }
}
\centering{
\subfloat[$p_y$ tracking using feedback linearization]{
    \includegraphics[width=0.22\textwidth]{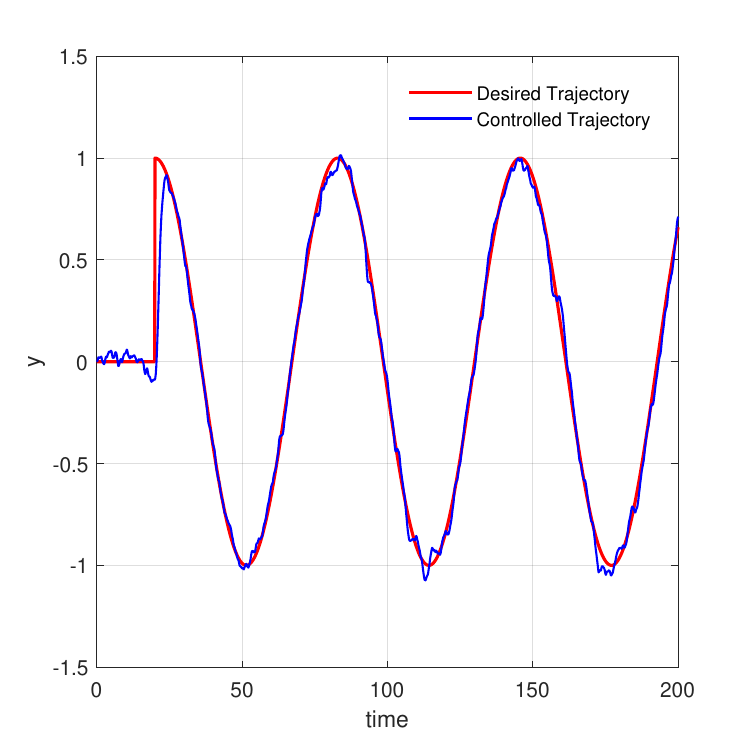}
    }
}
\centering{
\subfloat[$p_z$ stabilization using feedback linearization]{
    \includegraphics[width=0.22\textwidth]{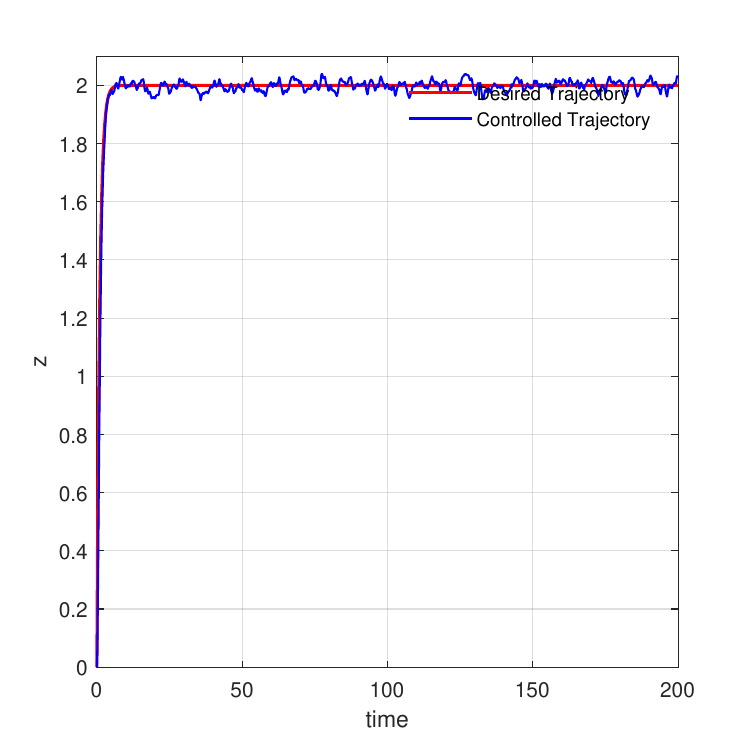}
    }
}
\centering{
\subfloat[3D trajectory tracking using feedback linearization]{
    \includegraphics[width=0.22\textwidth]{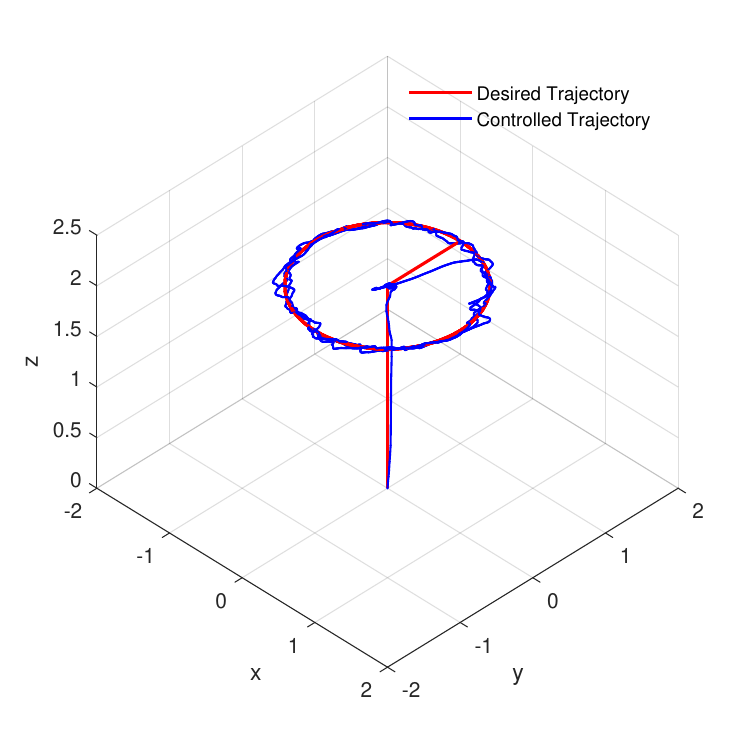}
    }
}
\caption{Trajectory tracking with process noise using feedback linearization in x,y direction while maintaining a constant altitude.}
\label{fig:FL-noise-tracking}
\end{figure}

\begin{figure}[h!]
\centering{
\subfloat[$p_x$ tracking using CLF-QP]{
    \includegraphics[width=0.22\textwidth]{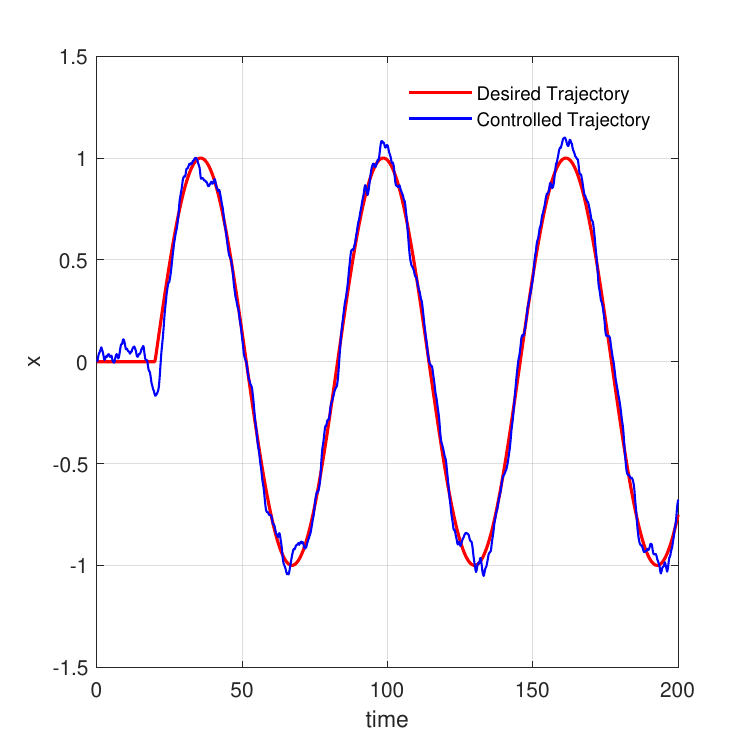}
    }
}
\centering{
\subfloat[$p_y$ tracking using CLF-QP]{
    \includegraphics[width=0.22\textwidth]{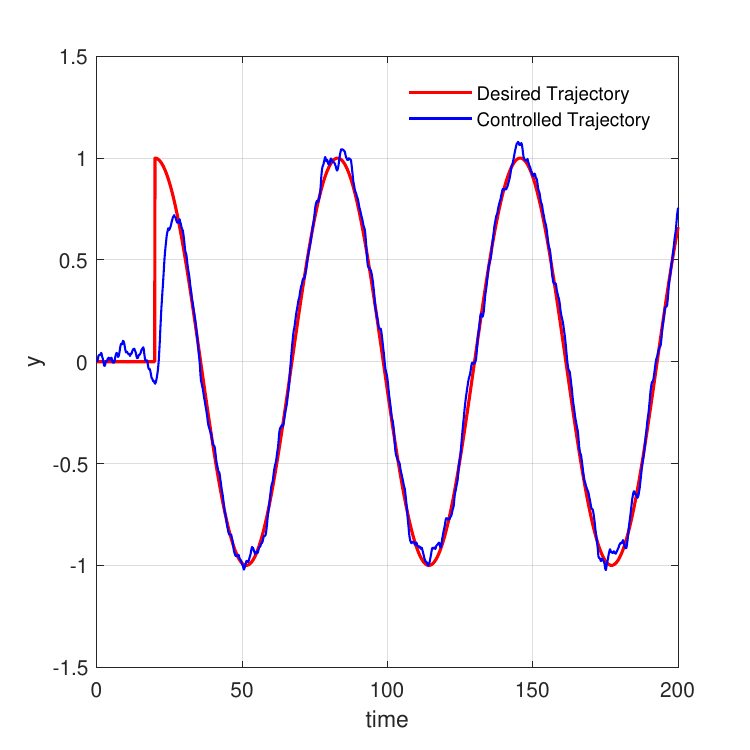}
    }
}
\centering{
\subfloat[$p_z$ stabilization using CLF-QP]{
    \includegraphics[width=0.22\textwidth]{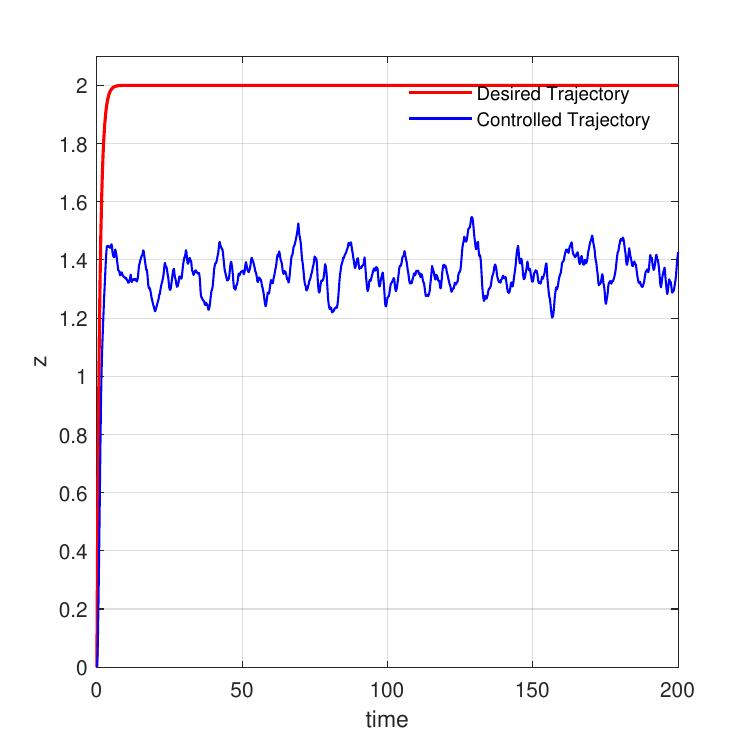}
    }
}
\centering{
\subfloat[3D trajectory tracking using CLF-QP]{
    \includegraphics[width=0.22\textwidth]{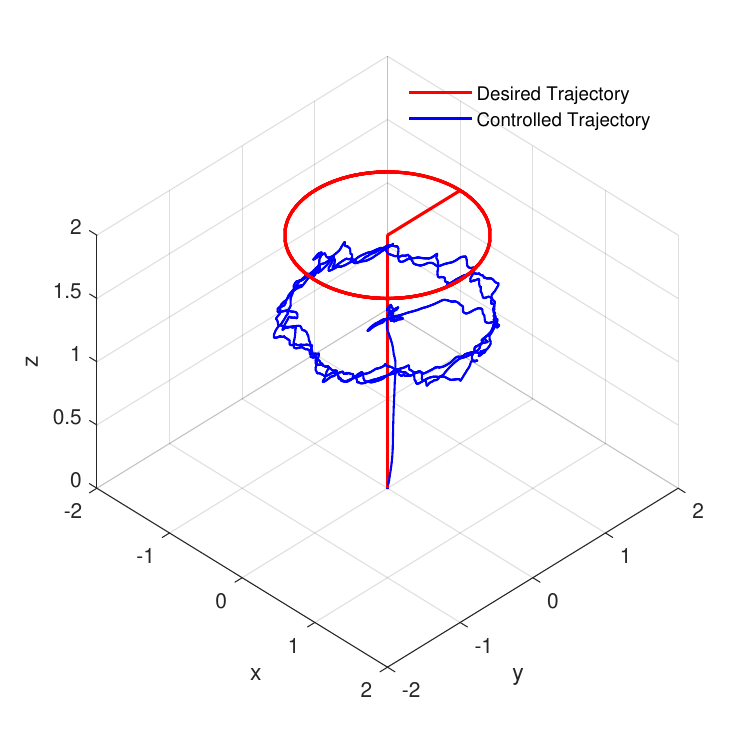}
    }
}
\caption{Trajectory tracking with process noise using CLF-QP in x,y direction while maintaining a constant altitude. The z direction does not stabilize to the desired height despite tuning the controller gain.}
\label{fig:CLF-QP-noise-tracking}
\end{figure}

This section includes results when using the controllers designed in section~\ref{sec:control_design_quad} to track the a trajectory when dynamics has an additive noise. In the plots, Figs.~\ref{fig:FL-noise-tracking},\ref{fig:CLF-QP-noise-tracking} it can be observed that feedback linearization performs better than CLF-QP. There could multiple reasons for this, 1) the domain of attraction due to chosen $P$ and $Q$ matrices for CLF might be small. To investigate this, we have used multiple $P$ and $Q$ matrices to increase the domain of attraction which led to blow up of optimization problem with no appreciable improvement in tracking. 2) The gains of the desired position to desired Euler angle computation in section~\ref{subsubsec:force-allocation} might be low. Increasing this did not improve the tracking. 3) To achieve tracking we need to numerically differentiate computed desired Euler angles twice. This differentiation along with low gain control might be a reason. Rapidly exponential CLF can be used to investigate this phenomenon, which was not performed here.    

\subsection{Quadrotor \& Pendulum}
\subsubsection{$\mathbf{y}_p = [a,b]^\top$}

\begin{figure}[t!]
\centering{
\subfloat[Pendulum position with $\xi$ controller]{
    \includegraphics[width=0.22\textwidth]{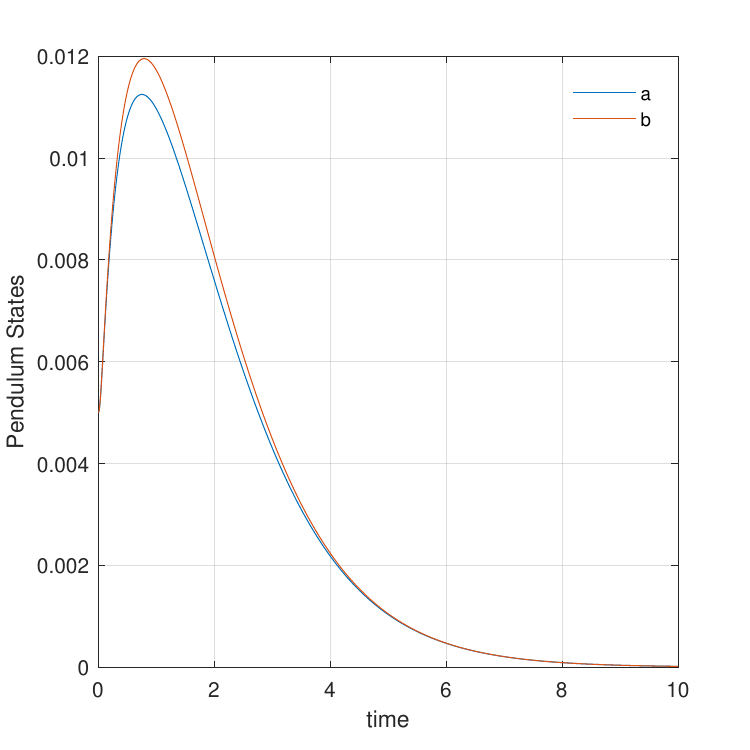}
    }
}
\centering{
\subfloat[Pendulum positions with $\xi^\prime$ controller]{
    \includegraphics[width=0.22\textwidth]{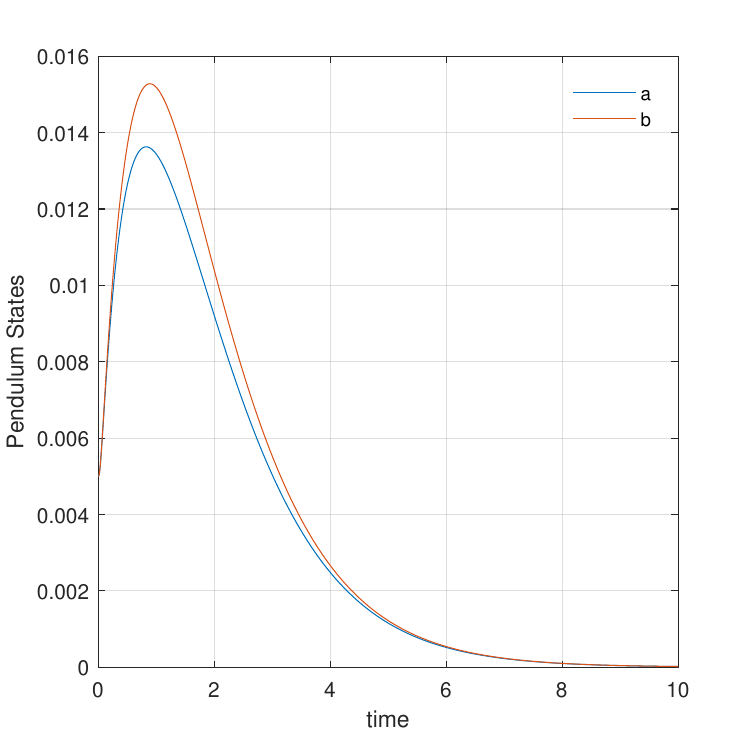}
    }
}
\centering{
\subfloat[Quadrotor accelerations with $\xi$ controller]{
    \includegraphics[width=0.22\textwidth]{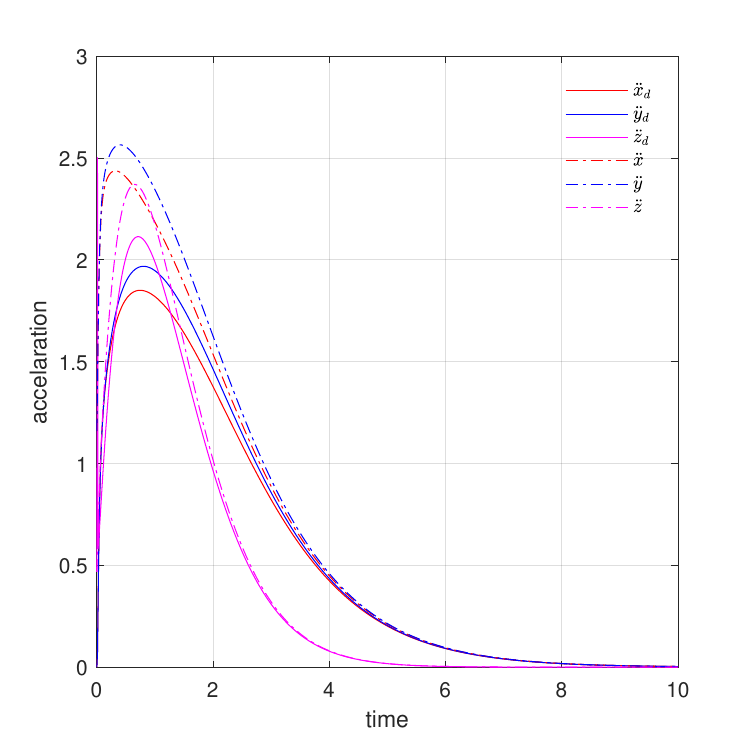}
    }
}
\centering{
\subfloat[Quadrotor accelerations with $\xi^\prime$ controller]{
    \includegraphics[width=0.22\textwidth]{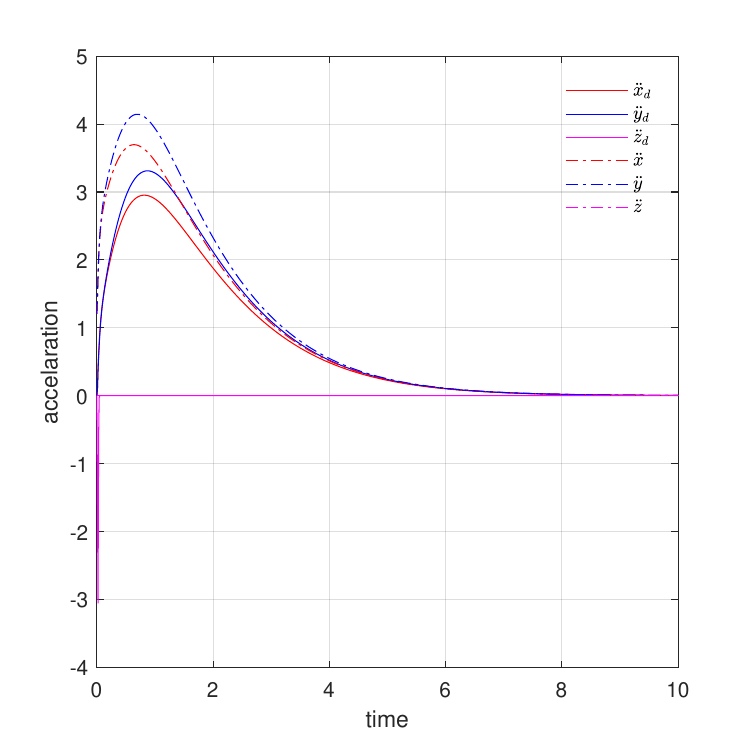}
    }
}
\caption{Balancing a inverted pendulum from initial disturbance with $\xi$ and $\xi^\prime$ controllers. Convergence is shown on pendulum position which aided by faster tracking of desired quadrotor accelerations}
\label{fig:pend_balance}
\end{figure}

\begin{figure}[t!]
\centering{
\subfloat[Pendulum position trajectory with $\xi$ controller]{
    \includegraphics[width=0.22\textwidth]{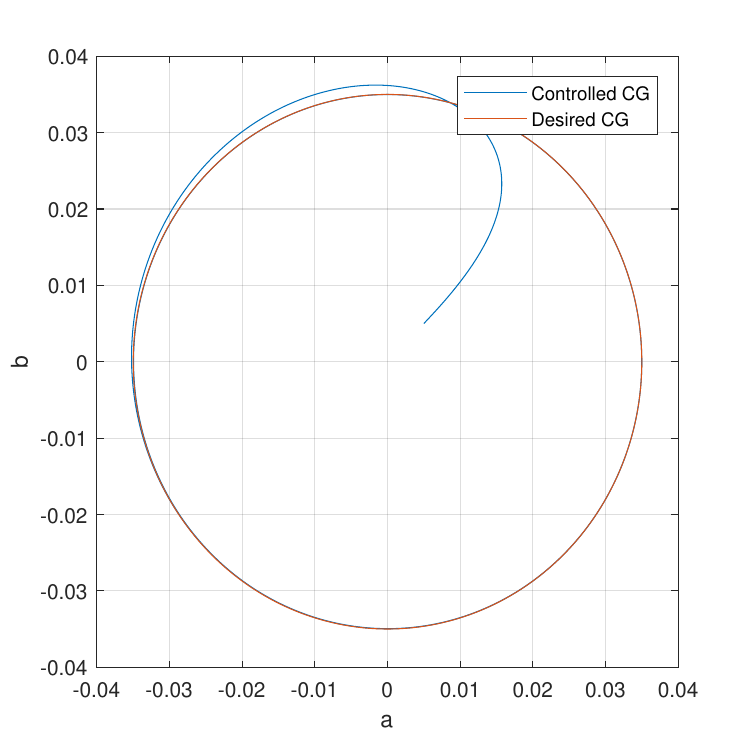}
    }
}
\centering{
\subfloat[Pendulum positions trajectory with $\xi^\prime$ controller]{
    \includegraphics[width=0.22\textwidth]{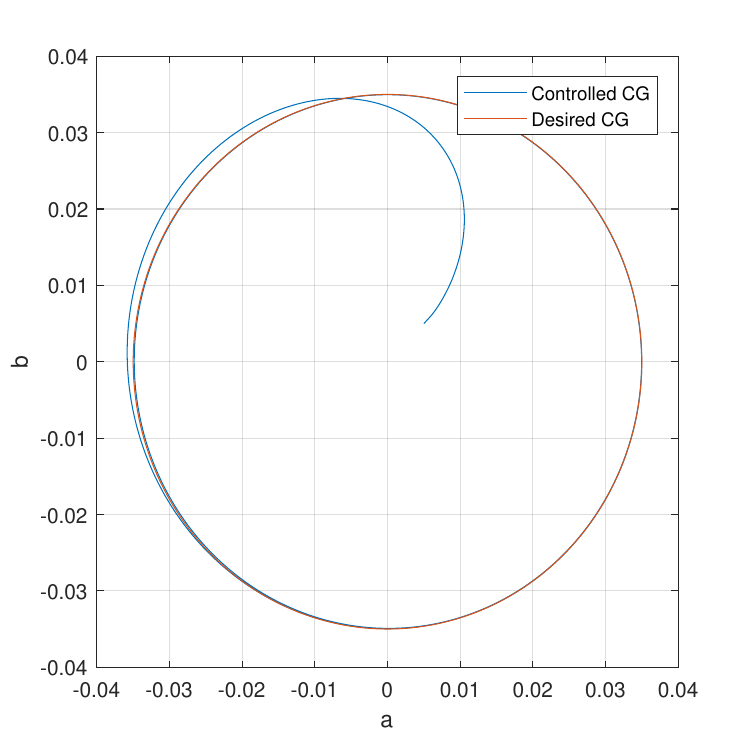}
    }
}
\caption{Swinging a inverted pendulum from initial disturbance to a circular trajectory with $\xi$ and $\xi^\prime$ controllers. Convergence to desired trajectory is achieved}
\label{fig:pend_tracking}
\end{figure}

Output tracking of pendulum position is first shown using the two controllers $\xi$ and $\xi^\prime$ described in Sect.~\ref{sec:fbklin_pend}. In Fig.~\ref{fig:pend_balance}, the pendulum position is initially perturbed away from the equilibrium. It can be seen that $[a,b]$ first grow in magnitude until convergence is achieved on quadrotor accelerations. Comparing $\xi$ and $\xi^\prime$ controllers, it can be seen that the overshoot of $[a,b]$ and commanded accelerations $\ddot{\mathbf{p}}$ are both smaller in magnitude for $\xi$. This is expected since additional effort from $\ddot{p_Z}$ can further help the system to stabilize. Trajectory tracking can be implemented as well shown in Fig.~\ref{fig:pend_tracking}. Here $[a,b]$ are driven to a fast circular trajectory at $0.1 \mathrm{Hz}$.

Although the result looks promising, it's not surprising that the zero dynamics of $[p_X,p_Y]^\top$ for $\xi^\prime$ and $[p_X,p_Y,p_Z]^\top$ for $\xi$ are only marginally stable, resulting them to go to $\infty$ as time goes on. 
\subsubsection{$\mathbf{y}_p = [a,b,p_X,p_Y]^\top$}
To address this problem of zero dynamics drifitng, we now introduce the result by employing the LQR controller method described in Sect.~\ref{sec:lqr_pend}. It is straightforward to apply this controller since the desired Euler angles $\mathbf{q}_d$ are directly fed into the same attitude controller designed in Sect.~\ref{subsec:Quad_Control}. Thus in principle, fast convergence of $\mathbf{q} \to \mathbf{q}_d$ will guarantee good tracking performance of $\mathbf{y}_p = [a,b,p_X,p_Y]^\top]$.
\begin{figure}[t!]
\centering{
\subfloat[Quadrotor position set-point]{
    \includegraphics[width=0.22\textwidth]{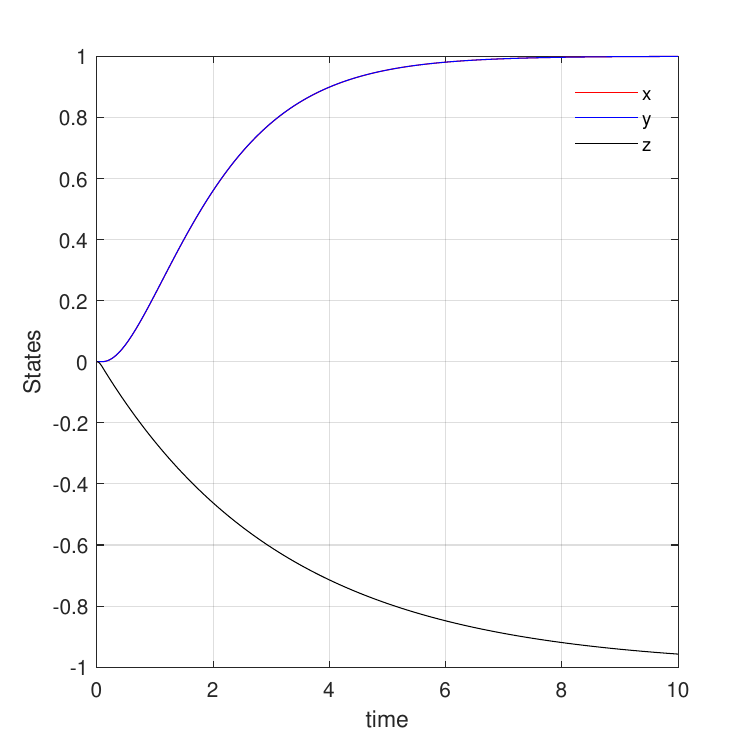}
    }
}
\centering{
\subfloat[Pendulum positions set-point]{
    \includegraphics[width=0.22\textwidth]{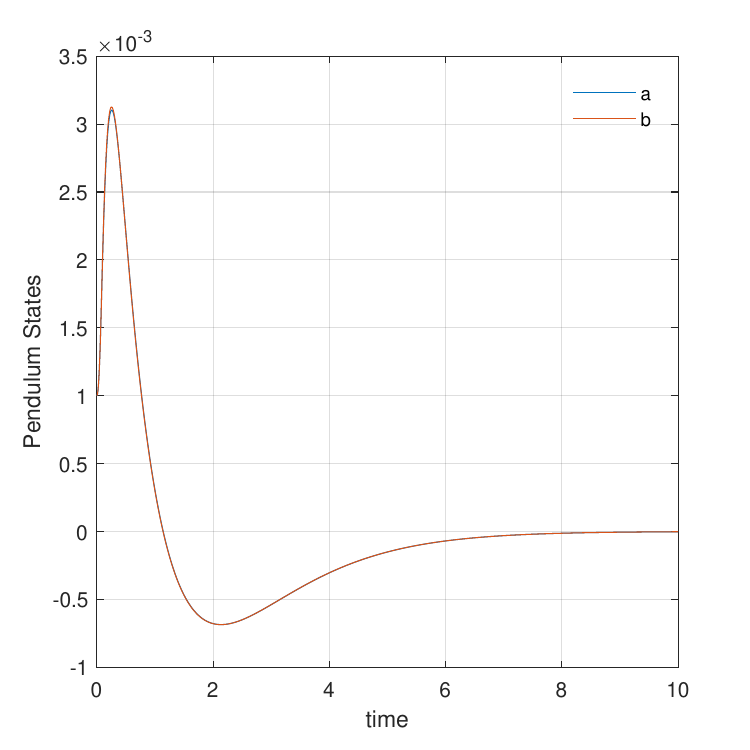}
    \label{fig:pend_quad_balance_ab}
    }
}
\caption{Moving the quadrotor to a position set-point while balancing the pendulum:$a_d = 0$; $b_d = 0$; $p_{Xd}(t) = 1$; $p_{Yd}(t) = 1$; $p_{Zd}(t) = -1$}
\label{fig:pend_quad_balance}
\end{figure}
Fig.~\ref{fig:pend_quad_balance} illustrates the pendulum balancing ability of the controller while driving the quadrotor position to a set-point. Beyond obvious convergence, it can also be seen that $[a,b]^\top$ now overshoots $0$ seen in Fig.~\ref{fig:pend_quad_balance_ab}, which is different from a pure exponential decay from Fig.~\ref{fig:pend_balance}.
\begin{figure}[t!]
\centering{
\subfloat[Quadrotor position trajectory tracking]{
    \includegraphics[width=0.22\textwidth]{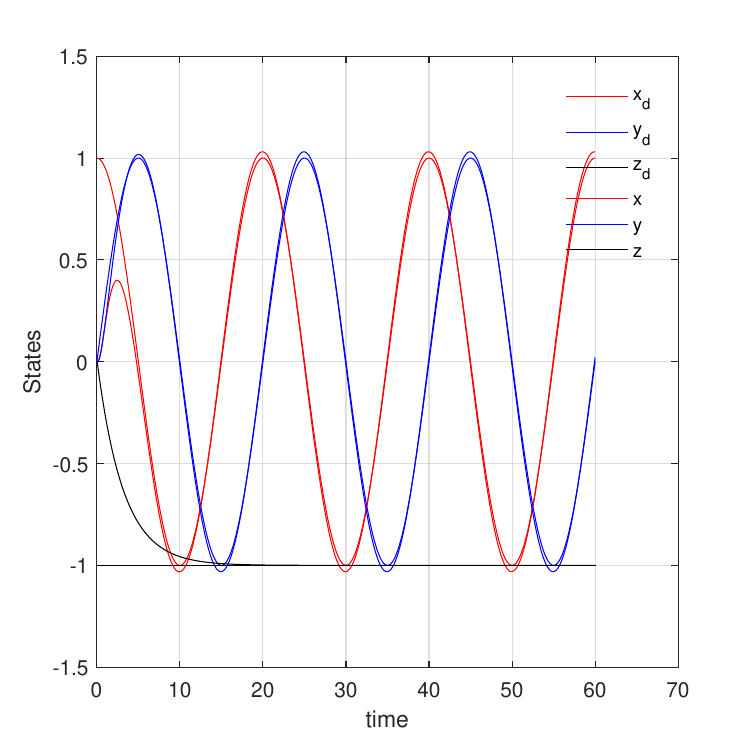}
    }
}
\centering{
\subfloat[Pendulum positions trajectory evolution]{
    \includegraphics[width=0.22\textwidth]{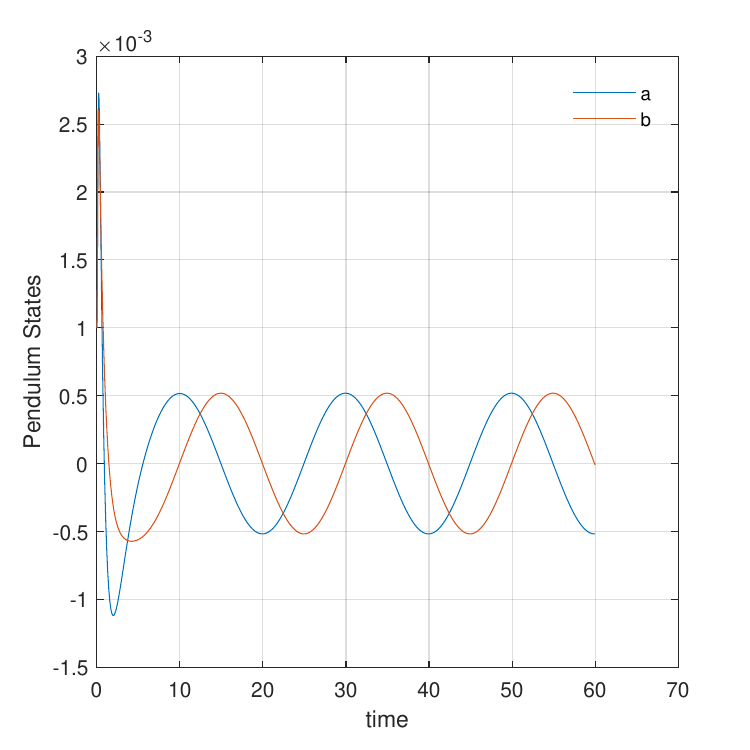}
    }
}
\caption{Tracking a quadrotor position trajectory while balancing the pendulum: $a_d = 0$; $b_d = 0$; $p_{Xd}(t) = R\cos{kt}$; $p_{Yd}(t) = R\sin{kt}$; $p_{Zd}(t) = -1$}.
\label{fig:pend_quad_tracking}
\end{figure}
This method also achieves good trajectory tracking performance when the maneuver is not aggressive (i.e closer to nominal). As shown in Fig.~\ref{fig:pend_quad_tracking}, rapid convergence of $\mathbf{p} \to \mathbf{p}_d(t)$ accompanies a bounded $\lVert [a,b]^\top \rVert$.

\section{CONCLUSION}
In this report, we employed and combined different types of linear and nonlinear controllers to achieve position tracking of combined quadrotor and pendulum systems. We use feedback linearization idea and CLF-QP for lower level attitude and altitude controller, as well as pendulumn position controller. To bridge between desired acceleration and desired attitude, we used a simple geometric transformation common in the quadrotor research community~\cite{mellinger2011minimum}. To achieve simultaneoud position tracking for both quadrotor and pendulum, we turn to linear-quadratic regulator to get reasonable performance under mild maneuvers. Multi-objective methods were attempted but not successful. Further improvements can be achieved by utilizing online or offline numerical optimization to generate feasible and stable trajectory for all relevant states.

The videos for results in Sect.~\ref{sec:simulation_results} are contained in the following link:
\url{https://caltech.box.com/s/ixkoob7m4l5aq6fmu3y5kasb0kysoa0z}







\bibliographystyle{IEEEtran}
\bibliography{IEEEabrv,references}

\end{document}